\documentclass[runningheads]{llncs}
\usepackage[T1]{fontenc}
\usepackage{graphicx}
\usepackage{url}
\begin{document}
\title{AI Agent-Driven Framework for Automated Product Knowledge Graph Construction in E-Commerce}

\author{Dimitar Peshevski \and Riste Stojanov \and Dimitar Trajanov}
\authorrunning{D. Peshevski and D. Trajanov}

\institute{Faculty of Computer Science and Engineering, \\ Ss. Cyril and Methodius University, \\ Skopje, N. Macedonia \\
\email{\{dimitar.peshevski,riste.stojanov,dimitar.trajanov\}@finki.ukim.mk}}

\maketitle

\begin{abstract}

The rapid expansion of e-commerce platforms generates vast amounts of unstructured product data, creating significant challenges for information retrieval, recommendation systems, and data analytics.
Knowledge Graphs (KGs) offer a structured, interpretable format to organize such data, yet constructing product-specific KGs remains a complex and manual process. This paper introduces a fully automated, AI agent-driven framework for constructing product knowledge graphs directly from unstructured product descriptions. Leveraging Large Language Models (LLMs), our method operates in three stages using dedicated agents: ontology creation and expansion, ontology refinement, and knowledge graph population. This agent-based approach ensures semantic coherence, scalability, and high-quality output without relying on predefined schemas or handcrafted extraction rules. We evaluate the system on a real-world dataset of air conditioner product descriptions, demonstrating strong performance in both ontology generation and KG population. The framework achieves over 97\% property coverage and minimal redundancy, validating its effectiveness and practical applicability. Our work highlights the potential of LLMs to automate structured knowledge extraction in retail, providing a scalable path toward intelligent product data integration and utilization.

\end{abstract}

\section{Introduction}

The rapid expansion of e-commerce and retail platforms has generated vast amounts of unstructured product data in the form of product descriptions, specifications, and user reviews. Structuring this information into machine-readable knowledge is essential for enabling advanced applications such as product recommendation, comparison, search, and analytics. Knowledge graphs (KGs), which represent entities and their relationships in a graph structure, have emerged as a powerful tool for organizing and integrating heterogeneous data sources. However, constructing product knowledge graphs at scale remains a labor-intensive and domain-specific task, often requiring extensive manual effort in defining ontologies and extracting relevant information.

This paper proposes an automated framework, driven by AI agents, for constructing knowledge graphs specifically tailored to product domains. Our approach leverages Large Language Models (LLMs) to automate the creation of fine-grained product ontologies and the subsequent population of KGs directly from product description corpora. The framework operates in three main stages: ontology creation and expansion, ontology refinement, and KG population, eliminating the need for manually crafted ontologies or extraction rules. We evaluate the framework using a real-world retail dataset and demonstrate its effectiveness. Our work contributes a scalable and practical solution for automated knowledge extraction in retail and e-commerce settings.

\section{Related Work}

The construction of product knowledge graphs, especially in e-commerce, has drawn significant interest from both academia and industry. This section surveys prior work across four categories: (1) Automated LLM-driven KG Construction, (2) Prompt-based Incremental KG Construction, (3) E-commerce Product KGs and Applications, and (4) Surveys, Roadmaps, and Practical Integrations.

\subsection{Automated LLM-based Knowledge Graph Construction}

Advances in LLMs such as GPT-4 and code models have enabled more scalable, less manually intensive KG construction pipelines.

Several efforts target LLM-based automation. Zhang et al. \cite{zhang2024extract} propose the Extract-Define-Canonicalize (EDC) framework, which separates open information extraction, schema creation, and canonicalization. Kommineni et al. \cite{kommineni2024human} aim to minimize human intervention with a pipeline incorporating competency-question-based ontology generation and LLM-driven evaluation. Zhu et al. \cite{zhu2024llms} evaluate GPT-4’s capabilities in KG construction and introduce a multi-agent AutoKG framework combining LLMs with external data.

To support topic-agnostic, fully automated pipelines, Lairgi et al. \cite{lairgi2024itext2kg} propose iText2KG—a zero-shot, plug-and-play system that distills unstructured input into KGs with minimal post-processing. Grapher \cite{melnyk2021grapher} generates end-to-end KGs by dividing tasks into entity and relation extraction using pre-trained LMs.

Bi et al. \cite{bi2024codekgc} present CodeKGC, which employs code LMs and schema-aware prompts for generative KG tasks, bolstered by rationale-based generation. Ning et al. \cite{ning2022knowledge} explore factual knowledge extraction via prompt templates and parameter tuning to improve accuracy.

\subsection{Prompt-based Incremental Knowledge Graph Construction and Reasoning}

Prompt learning plays a key role in cross-domain KG tasks and reasoning. Cui et al. \cite{cui2024prompt} propose KG-ICL, combining in-context learning with a prompt graph and unified tokenization, achieving strong generalization across 43 KGs. Xie et al. \cite{xie2022promptkg} introduce PromptKG, a toolkit integrating prompt-learning methods for KG applications.

For KG completion, Yao et al. \cite{yao2025exploring} present KG-LLM, which uses entity/relation prompts for triple classification and relation prediction, finding smaller tuned models can outperform larger LLMs like GPT-4 in some cases.

\subsection{E-Commerce Product Knowledge Graphs: Construction and Use Cases}

E-commerce KGs face unique structural and scalability challenges, leading to domain-specific methodologies and applications.

Xu et al. \cite{xu2020product} propose an embedding method tailored to product KGs using multimodal data for KG completion and recommendation. Zalmout et al. \cite{zalmout2021all} present a roadmap of best practices for scalable e-commerce KG construction.

Large-scale efforts include Amazon’s Product Graph \cite{dong2018challenges}, which uses annotation, distant supervision, and attribute extraction, and AliMeKG \cite{li2020alimekg}, a domain KG supporting customer-facing tasks at Alibaba.

Other works focus on KG generation from unstructured text, such as extracting product facts from Q\&A data \cite{sant2020generating}, and converting dialogues into triples for recommendations \cite{regino2023leveraging,regino2022natural}. Perrot et al. \cite{perrot2024knowledge} explore retail data integration via semantic web and KG techniques in a real-world case study.

\subsection{Surveys, Frameworks, and Forward-looking Perspectives}

Recent surveys synthesize LLM-KG developments and outline future directions. Pan et al. \cite{pan2024unifying} categorize work into KG-enhanced LLMs, LLM-augmented KGs, and synergistic frameworks for bidirectional reasoning. Liang et al. \cite{liang2024survey} review LLM-based KG applications in complex product design, highlighting technical patterns and domain-specific challenges.

\section{Methodology}

\begin{figure}[h]
\centering
\includegraphics[width=1.0\textwidth]{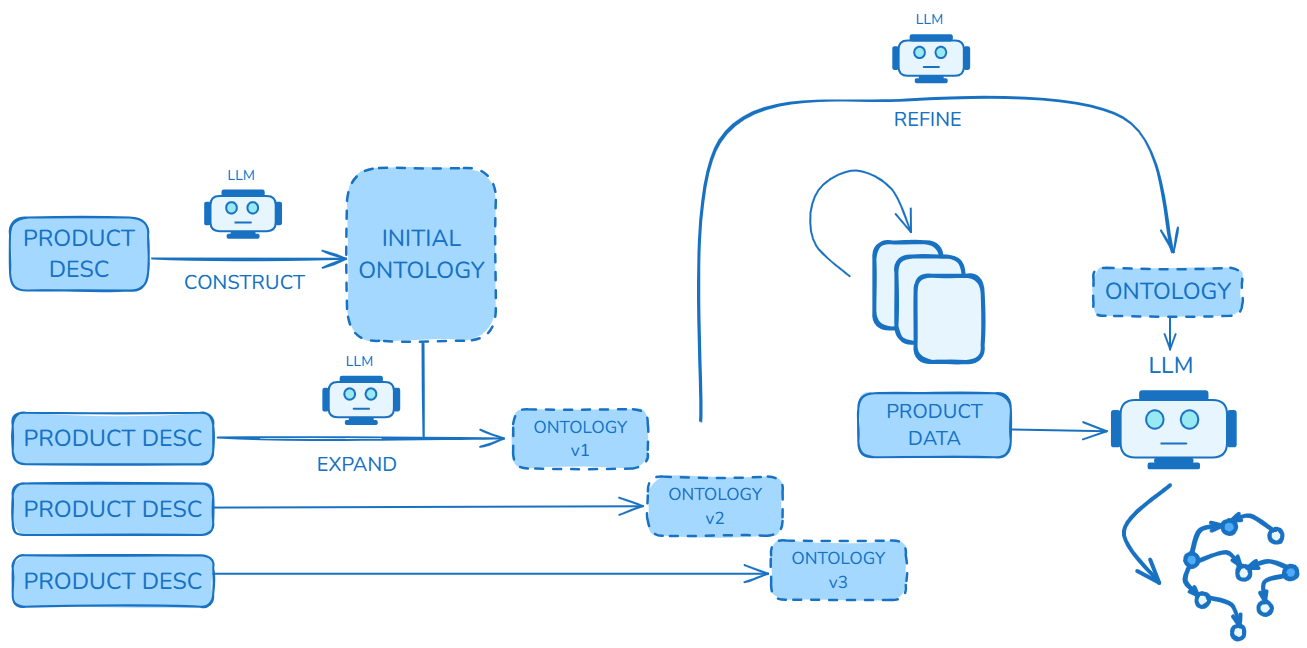}
\caption{Agent-based workflow consisting of ontology creation, refinement, and KG population. Each step is handled by a dedicated LLM-powered agent operating within a modular pipeline.}
\label{pipeline}
\end{figure}

Our framework consists of three main stages: (1) ontology creation and expansion, (2) ontology refinement, and (3) knowledge graph population. Each stage is orchestrated through a modular pipeline of LLM-powered agents designed to balance automation with fine-grained control. This section describes the agent-based workflow and provides implementation details. Figure \ref{pipeline} illustrates the overall architecture.

\subsection{The Agent-based Workflow}

\subsubsection{Ontology Creation and Expansion}

We begin by sampling representative product descriptions from the corpus, focusing on coverage across product categories. An LLM-based agent is employed to extract initial ontology elements. The agent identifies product classes, attributes, and relationships, organizing them into RDF/Turtle format with clearly defined \texttt{rdfs:domain}, \texttt{rdfs:range}, and descriptive \texttt{rdfs:comment} annotations. This ensures that each class and property is semantically interpretable and machine-readable. Ontology expansion proceeds iteratively. We present additional product descriptions to the agent using a prompt that instructs it to generalize beyond individual instances and extend the schema where necessary. The agent integrates new classes or properties discovered in these samples without removing existing elements, preserving schema stability. Attribute descriptions specify expected value formats and units, which are critical for ensuring consistency during the KG population. We typically iterate this expansion process over approximately 30 product samples per category until the number of new ontology elements added per iteration significantly diminishes (indicating a plateau), balancing coverage with schema manageability.

\subsubsection{Ontology Refinement}

While the initial ontology captures local patterns, it may lack generality, consistency, or optimal modularity. To address this, we perform zero-shot refinement leveraging the LLM's encoded knowledge. We provide the complete ontology as input, prompting the LLM to suggest revisions, merges, or extensions aimed at improving generality, reducing redundancy, clarifying ambiguities, and enhancing adaptability to diverse product domains. This step often leads to splitting compound attributes into separate properties, tightening domain and range restrictions, or enhancing attribute comments for greater precision.

\subsubsection{Knowledge Graph Population}

In the final step, we populate the knowledge graph with instance-level data extracted from the product corpus. Given the refined ontology and individual product descriptions, the LLM-based agent generates RDF triples that capture specific product details. The agent is guided by a prompt that instructs it to map product attributes and relationships precisely according to the ontology, translating values into English where necessary and respecting the expected units and formats defined in the \texttt{rdfs:comments}. Triples are only generated when values are explicitly available in the description, avoiding hallucination or inference beyond the provided data. The generated triples are compiled into a unified knowledge graph, enabling downstream applications such as product search, recommendation, and analytics.

\subsection{Implementation Details}

We implemented the framework using the ChatGPT 4.1 Mini model, chosen for its favorable balance of cost and performance. The overall pipeline is automated using a combination of Python scripts, RDF libraries (such as RDFLib), and a lightweight orchestration layer that manages sampling, prompt generation, LLM invocation, and output integration. The prompts used at each stage are designed to be modular and composable, enabling iterative improvements and rapid adaptation to new domains or product types.
\subsection{Evaluation}

We evaluate our framework on a real-world dataset from a retail store, consisting of 291 product descriptions for the category "air conditioners".

The evaluation is conducted along three dimensions:

\begin{itemize}
    \item Ontology Coverage: We assess the completeness of the automatically generated ontologies by measuring the number of extracted classes, attributes, and relationships.

    \item Ontology Quality: We perform qualitative evaluation of the ontology’s coherence, generality, and usability, with expert annotation to identify redundancies, inconsistencies, or missing elements.

    \item Knowledge Graph Population: We automatically evaluate the populated knowledge graph by measuring the number of generated RDF triples and the proportion of ontology properties that are instantiated in the graph.
\end{itemize}

\section{Results}

The constructed ontology comprises 42 classes and 69 properties, including 20 data attributes and 49 object relationships. This reflects a high level of coverage across the air conditioners product category. We manually reviewed the schema and found it to be modular, hierarchically structured, and readily extendable, allowing for seamless integration of new products and attributes as needed.

For ontology quality, we conduct a qualitative assessment involving manual annotation. We manually evaluated the ontology’s coherence, generality, and usability, focusing on aspects such as redundancy, consistency, and completeness. We report minimal redundancy and no significant inconsistencies, confirming that the ontology generalizes well across the product category while maintaining sufficient specificity for downstream tasks.

Regarding knowledge graph population, the framework successfully processed 282 out of 291 product descriptions, with failures on only nine instances (3\%) due to invalid RDF outputs. The resulting knowledge graph contains 7,459 RDF triples, achieving 97.1\% coverage of the properties defined in the ontology. This indicates that nearly all classes and attributes were populated with real product data, demonstrating the system’s robustness and practical applicability.

Overall, the results demonstrate that the proposed framework produces high-coverage, fine-grained ontologies and achieves strong performance in knowledge graph population. These outcomes highlight the framework’s effectiveness and practicality for large-scale retail applications, enabling automated, scalable, and accurate product knowledge extraction.

\section{Conclusion and Future Work}

In this work, we introduce an AI agent–driven framework for fully automated knowledge graph construction from product descriptions. Our framework leverages the capabilities of large language models to perform end-to-end ontology creation, iterative refinement, and knowledge graph population, enabling the scalable extraction of structured product knowledge. By eliminating the need for manual schema design, rule engineering, or annotated datasets, the framework significantly reduces development overhead and improves adaptability to new domains and product types.

Looking ahead, several technical extensions could further strengthen the framework. Integrating multimodal data, such as product images, specifications sheets, and customer reviews, can improve coverage and semantic depth. Enhancing population accuracy may be achieved through fine-tuned LLMs or hybrid symbolic-neural pipelines. Expanding to domains like finance or healthcare will require addressing domain-specific ontologies and compliance needs. Finally, adding continuous updates from streaming product data will support dynamic, real-time KG construction for retail and e-commerce applications.

Overall, the proposed framework provides a robust and extensible foundation for next-generation product intelligence systems, supporting advanced downstream applications such as recommendation engines, search optimization, and automated catalog enrichment.

\begin{credits}
\subsubsection{\ackname} This publication is based upon work from COST Action CA23147 GOBLIN - Global Network on Large-Scale, Cross-domain and Multilingual Open Knowledge Graphs, supported by COST (European Cooperation in Science and Technology, \url{https://www.cost.eu}).
\end{credits}

\bibliographystyle{splncs04}
\bibliography{references}

\end{document}